\documentclass{article}
\usepackage{spconf,amsmath,graphicx}
\usepackage{amssymb}
\usepackage{multirow}
\usepackage{graphicx}
\usepackage{booktabs}

\title{M$^3$TN: Multi-gate Mixture-of-Experts based Multi-valued Treatment Network for Uplift Modeling}
\name{Zexu Sun$^{\dagger,\ddagger}$ \qquad Xu Chen$^{\dagger,\ddagger,*}$\thanks{* Corresponding author}}

\address{$^{\dagger}$ Gaoling School of Artificial Intelligence, Renmin University of China, Beijing, China\\
$^{\ddagger}$ Beijing Key Laboratory of Big Data Management and Analysis Methods
\\
\texttt{\{sunzexu21, xu.chen\}@ruc.edu.cn}
}
\begin{document}
\ninept  
\maketitle
\begin{abstract}
Uplift modeling is a technique used to predict the effect of a treatment (\textit{e.g.}, discounts) on an individual's response. Although several methods have been proposed for multi-valued treatment, they are extended from binary treatment methods. There are still some limitations.
Firstly, existing methods calculate uplift based on predicted responses, which may not guarantee a consistent uplift distribution between treatment and control groups. Moreover, this may cause cumulative errors for multi-valued treatment.
Secondly, the model parameters become numerous with many prediction heads, leading to reduced efficiency.
To address these issues, we propose a novel \underline{M}ulti-gate \underline{M}ixture-of-Experts based \underline{M}ulti-valued \underline{T}reatment \underline{N}etwork (M$^3$TN). M$^3$TN consists of two components: 1) a feature representation module with Multi-gate Mixture-of-Experts to improve the efficiency; 2) a reparameterization module by modeling uplift explicitly to improve the effectiveness. We also conduct extensive experiments to demonstrate the effectiveness and efficiency of our M$^3$TN.

\end{abstract}

\section{Introduction}
\label{sec:intro}

An important aim of the online platform is to enhance user engagement and increase platform revenue through effective online marketing strategies. These strategies involve implementing well-designed incentives, \textit{e.g.} coupons, discounts, and bonuses~\cite{liu2023explicit}. 
For a successful online marketing strategy to achieve effective delivery and minimize costs, it is vital to accurately identify the target user group for each incentive.
Regarding this as a causal inference problem, as a result, it is imperative to estimate the individual treatment effect (ITE)~\cite{zhang2021unified,sun2023offline} (also known as uplift). 
In practice, it is common to observe only one type of user response, which can be attributed either to a specific incentive (\textit{i.e}., treatment group) or the absence of an incentive (\textit{i.e}., control group). 
Therefore, uplift modeling is proposed to solve this problem, and many previous works have verified its effectiveness in online marketing~\cite{kawanaka2019uplift,sun2023robustness}. 

The existing uplift modeling works mainly focus on three research lines. 1) \textit{Meta learner-based}. The basic idea of this line is to use any off-the-shell estimator as the base learner. 
The two most representative models of this line are S-Learner~\cite{kunzel2019metalearners} and T-Learner~\cite{kunzel2019metalearners}, which use one global estimator or estimators corresponding to the treatment types. There are also other meta-learners that introduce other structures, such as X-Learner~\cite{kunzel2019metalearners}, R-Learner~\cite{nie2021quasi}, etc.
2) \textit{Tree (or Forest)-based}. The basic idea of this line is to divide users into different subgroups in the feature space. 
By designing the diverse splitting criterions~\cite{rzepakowski2010decision,zhao2017uplift}, the uplift can be predicted on each leaf node. Especially, Causal Forest \cite{wager2018estimation} ensembles multiple trees to estimate the heterogeneity. 
And based on this, some works are proposed for uplift modeling with non-binary treatment~\cite{lechner2018modified,wan2022gcf}.
3) \textit{Neural network-based}. The basic idea of this line is to leverage the neural network with more flexible structures to predict the uplift. 
This line aims to leverage the neural network with more flexible structures to predict the uplift. 
Additionally, there are works based on representation learning, such as TARNet~\cite{shalit2017estimating}, CFRNet~\cite{shalit2017estimating}, Dragonnet~\cite{shi2019adapting}, etc., to mitigate biases in the dataset.
In this paper, we focus on a neural network-based line because the methods of this line have greater flexibility and generalizability compared to the methods of other lines.

Although the existing uplift models have shown promising results in the online marketing, most of them have focused on the setting where the treatment is binary.
However, in many real-world scenarios, the treatment may be multiple, such as the discount of an online platform. There are few works that focus on the multi-valued treatment uplift modeling. 
In recent years, although there some works proposed for this setting (\textit{e.g.}. MEMENTO~\cite{mondal2022memento}, Hydranet~\cite{velasco2022hydranet}), they directly extend the above binary treatment methods (\textit{e.g.}, CFRNet~\cite{shalit2017estimating}, Dragonnet~\cite{shi2019adapting}), which leads to decreased model efficiency. 
Additionally, in view of the uplift problem, it is crucial that the predicted uplift is distributed equally between the treatment and control groups. Consequently, without explicitly modeling uplift, it is not possible to achieve this equilibrium, particularly when the number of treatment values increases, this may result in cumulative errors.


To alleviate the aforementioned problems, in this paper, we propose a novel uplift model \underline{M}ixture-of-Experts based \underline{M}ulti-valued \underline{T}reatment \underline{N}etwork or M$^3$TN for short.
Specially, our M$^3$TN contains two customized modules: 1) a feature representation network with Multi-gate Mixture-of-Experts to improve efficiency; 
2) a reparameterization module by modeling uplift explicitly to improve the effectiveness. 
Furthermore, we conduct experiments on both public and real-world production datasets to demonstrate the effectiveness and efficiency of our M$^3$TN. Especially, our contributions are summarized in the following:
\begin{itemize}
    \item We design a feature representation network with Multi-gate Mixture-of-Experts for model efficiency.
    \item We propose a reparameterization module by modeling uplift explicitly to improve the effectiveness of uplift prediction.
    \item We conduct extensive experiments on a public dataset and a real product dataset, and the experimental results demonstrate the effectiveness and efficiency of our M$^3$TN. 
\end{itemize}

\section{Preliminaries}
Let $\mathcal{D}=\{(\boldsymbol{x}_i,t_i,y_i)\}_{i=1}^n$ represent the observed dataset with $n$ samples. 
Without losing the generality, $\boldsymbol{x}_i \in \mathcal{X} \subset \mathbb{R}^d$ is a $d$ dimensional feature vector. 
$y_i \in \mathcal{Y}$ is the response variable, where $\mathcal{Y}$ can be in ether binary or continuous space. $t_i \in \mathcal{T} \in \{0, 1, \ldots , K\} (K\geq 2)$ is the treatment indicator variable with $K+1$ values, \textit{e.g.,} the different discounts. 

To formalize the problem, we follow the Neyman-Rubin potential outcome framework~\cite{rubin2005causal} to define the uplift modeling problem with multi-valued treatment.
Let $y_i(k)$ and $y_i(0)$ denote the potential outcomes of whether user $i$ receive an incentive $t_i = k \in \{1,\ldots,K\}$ or not treated. 
The incremental improvement caused by a specific treatment $k$ (individual treatment effect or uplift) is denoted by $\tau_i^k$, which is defined as:
\begin{equation}
    \tau_i^k = y_i(k)-y_i(0),
\end{equation}

Since we can only observe one response (\textit{i.e.} $y_i(k)$ or $y_i(0)$) of each user, we can not have the true uplift label $\tau_i^k$. 
Fortunately, with some adequate assumptions~\cite{zhang2021unified}, we can use the conditional average treatment effect (CATE) as an unbiased estimator for the uplift, where CATE is defined as:
\begin{equation}
\begin{aligned}
\tau_i^k(\boldsymbol{x}_i) & =\mathbb{E}(y_i(k)-y_i(0)|\boldsymbol{x}_i) \\
& =\underbrace{\mathbb{E}(y_i(k) | t_i=k, \boldsymbol{x}_i)}_{\mu_k(\boldsymbol{x}_i)}-\underbrace{\mathbb{E}(y_i(0)| t_i=0, \boldsymbol{x}_i)}_{\mu_0(\boldsymbol{x}_i)} .
\end{aligned}
\end{equation}
which can be interpreted as the expected treatment effect between the $t_i=0$ and $t_i=k$ given $\boldsymbol{x}_i$.

For the brevity of notation, we will omit the subscript $i$ in the following if no ambiguity arises. 
Intuitively, the desired objective can be expressed as the difference between two conditional means $\hat{\tau}^k(\boldsymbol{x})=\mu_k(\boldsymbol{x})-\mu_0(\boldsymbol{x})$. 
After we obtain the uplift predictions $\hat{\tau}^k(\boldsymbol{x})$ for all $k$ different incentives, we can rank the users by using the evaluation metrics and make the final decision for the treatment assignment.

\section{Method}
\label{sec:format}
\begin{figure}[!t]
    \centering
    \includegraphics[width=\linewidth]{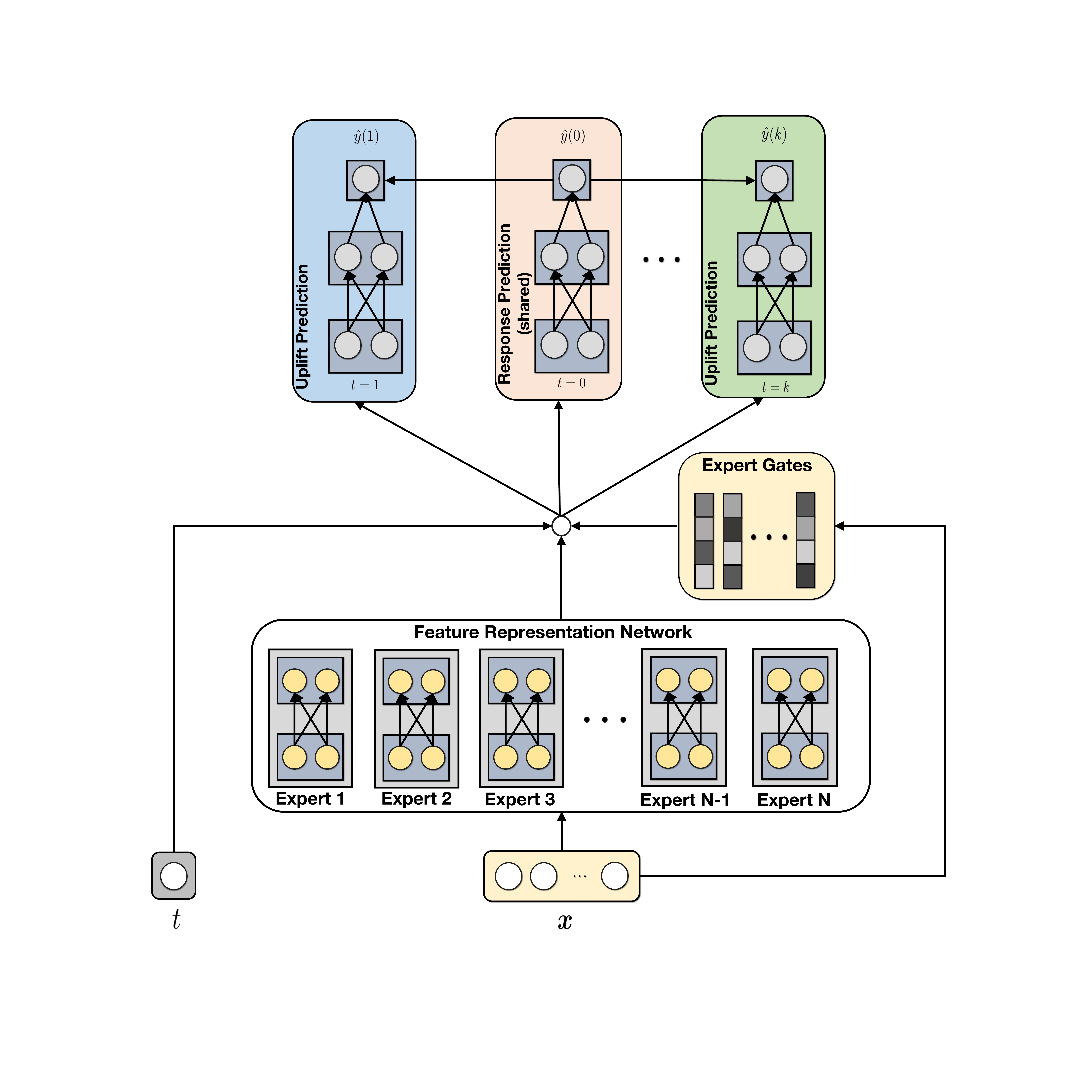}
    \caption{The architecture of the \underline{M}ixture-of-Experts based \underline{M}ulti-valued \underline{T}reatment \underline{N}etwork (M$^3$TN).}
    \label{fig:model}
\end{figure}
\subsection{Architecture}
\label{sec:arch}
The proposed \underline{M}ulti-gate \underline{M}ixture-of-Experts based \underline{M}ulti-valued \underline{T}reatment \underline{N}etwork (M$^3$TN), is shown in Fig.~\ref{fig:model}. Given a sample $\{\boldsymbol{x}, t, y\}$,  $\boldsymbol{x}$ is the features of the users, where categorical features are converted into embedding vectors. 

The feature representation module consists of $N$ expert networks, and for each prediction head, there is a gating network that ensembles the results from all experts. 
This module has fewer parameters compared to the shared bottom in MEMENTO~\cite{mondal2022memento} and Hydranet~\cite{velasco2022hydranet}, which can improve the efficiency even if the number of prediction heads increases.
Then, to predict the response, we use the separate prediction heads for treatment and control groups. 
Specially, a shared control response prediction head is used to predict $\mu_0(\boldsymbol{x})$, for the samples in the treatment groups, the prediction heads are used to model the uplift $\hat{\tau}^k(\boldsymbol{x})$.
The design of this reparameterization module can improve the effectiveness of the model by reducing the cumulative errors. 
Moreover, we use $h_0$ to denote the shared response prediction in the control group and use $h_1, h_2, \ldots, h_k$ to denote the uplift prediction heads for each treatment group. 
The parameters of the two modules are optimized jointly, and the objective of the model is defined as:
\begin{equation}
\resizebox{\linewidth}{!}{
$\min\limits_{\Theta}L_{M^3TN}  = \mathcal{L}(\mu_0(\boldsymbol{x}), y(0)) + \sum\limits_{k=1}^K \mathcal{L}(\mu_k(\boldsymbol{x}), y(k)) + \lambda \|\Theta\|_2.$}
    \label{eq:loss}
\end{equation}
where $\mu_k(\boldsymbol{x})=\mu_0(\boldsymbol{x})+\hat{\tau}^k(\boldsymbol{x})$, and $\hat{\tau}^k(\boldsymbol{x})$ is predicted by using the uplift prediction head $h_k$. $\mathcal{L}(\cdot, \cdot)$ represents the Mean Square Error (MSE) loss, $\lambda$ is to control the weight of model parameters regularization for the trade-off.

\subsection{Feature Representation Module}
To improve the efficiency and decrease the model parameters of uplift models with multi-valued treatment, we use a feature representation module with Multi-gating Mixture-of-Experts (MMoE).  
While MoE was first developed as an ensemble method of multiple individual models, Eigen \textit{et al}.~\cite{eigen2013learning} and Shazeer~\textit{et al}. \cite{shazeer2017outrageously} turn it into basic building blocks (MoE layer) and stack them in a neural network. The MoE layer has the same structure as the MoE model but
accepts the output of the previous layer as input and outputs to a successive layer. 
The whole model is then trained in an end-to-end way. 
To capture the sample differences between treatment and control groups with less model parameters compared to the shared-bottom multi-valued treatment uplift model, the Multi-gate MoE (MMoE) is introduced into the feature representation module. 
Then, the feature representation $\phi_k$ of each prediction head can be formulated as:
\begin{equation}
\phi_k(\boldsymbol{x}) =\sum_{n=1}^N g_k(\boldsymbol{x}) f_n(\boldsymbol{x}), \forall k \in\{0,1, \ldots, K\},
\end{equation}
where $f_n$ represents the expert layer, $g_k$ is the gating network for the prediction head $h_k$. 

Moreover, $f_n$ are implemented by multilayer perceptrons with ReLU activations. The gating networks $g_k$ are simply linear transformations of the input with a softmax layer:
\begin{equation}
g_k(\boldsymbol{x})=\operatorname{softmax}\left(\boldsymbol{W}_{g_k} \boldsymbol{x}\right).
\end{equation}
where $\boldsymbol{W}_{g_k} \in \mathbb{R}^{n \times d}$ is a trainable matrix, $n$ is the number of experts and $d$ is the feature dimension.

\subsection{Reparameterization Module}
For the uplift modeling problem, it is crucial that the distribution of the predicted uplift $\hat{\tau}^k(\boldsymbol{x})$ is consistent between the treatment and control groups. 
However, the existing multi-valued treatment uplift models build the response directly,
which is not possible to achieve this equilibrium, particularly when the number of treatment values increases, and this will lead to cumulative errors for treatment assignment.
To solve this, instead of estimating the user responses, we use a reparameterization module for building the uplift explicitly.
In particular, the control response prediction head $h_0$ is shared for all the samples. For other prediction heads $h_1, h_2, \ldots, h_k$, we use them to predict the uplift $\hat{\tau}^k(\boldsymbol{x})$ of different treatment assignments for a user. Then, the reparameterization module can be formulated as:
\begin{equation}
\begin{gathered}
\mu_0(\boldsymbol{x}) = h_0\left(\phi_0\right),\\
\hat{\tau}^k(\boldsymbol{x})=h_k\left(\phi_k\right), \forall k\in \{1, 2,\ldots, K\}.
\end{gathered}
\end{equation}
where $\mu_0(\boldsymbol{x})$ is the control response prediction, $\hat{\tau}^k(\boldsymbol{x})$ is the uplift prediction. Then, $\mu_k(\boldsymbol{x})$ can be estimated by an additive function $\mu_k(\boldsymbol{x})=\mu_0(\boldsymbol{x}) + \hat{\tau}^k(\boldsymbol{x})$.

By leveraging this module, the prediction head $h_k$ builds the uplift $\hat{\tau}^k(\boldsymbol{x})$ explicitly, and hard-codes that the shared structure between the responses of a user is additive. 
According to explicitly build the uplift, the reparameterization module can reduce the uplift prediction error.
To train our M$^3$TN, the two modules are jointly optimized via Eq.~\eqref{eq:loss}.

\section{Experiments}
In this section, we conduct experiments to answer the following research questions:
\begin{itemize}
    \item \textbf{RQ1}: How is the performance of our M$^3$TN compared with other baselines on both MultiTwins and Production datasets?
    \item \textbf{RQ2}: What is the role each module plays in our M$^3$TN?
    \item \textbf{RQ3}: What is the model efficiency of our M$^3$TN compared with other baselines?
    \item \textbf{RQ4}: How is the performance of our M$^3$TN sensitive to the hyperparameters?
\end{itemize}
\begin{table*}[!t]
    \centering
    \caption{Overall comparison between our M$^3$TN and the baselines on MultiTwins and Production datasets, where the best and second best results are marked in bold and underlined, respectively.}\label{tab:overall}
\resizebox{\linewidth}{!}{
    \begin{tabular}{cccccccccc}\toprule
      \multirow{2}{*}{\textbf{Methods}}   &\multicolumn{4}{c}{MultiTwins Dataset} & &\multicolumn{4}{c}{Production Dataset} \\\cline{2-5} \cline{7-10}
         & mQini$\uparrow$ & sdQini$\downarrow$& mKendall$\uparrow$ & sdKendall$\downarrow$ & & mQini$\uparrow$ & sdQini$\downarrow$ & mKendall$\uparrow$ & sdKendall$\downarrow$\\\midrule
S-Learner &0.5930 $\pm$ 0.0243 & 0.1833  $\pm$ 0.0466  &0.5033 $\pm$ 0.0322 &\underline{0.1814} $\pm$ 0.0271 &      &0.5807 $\pm$ 0.0397 &0.2025 $\pm$ 0.0377 &0.5833 $\pm$ 0.0176 &\underline{0.2133} $\pm$ 0.0156 \\
T-Learner &0.6586 $\pm$ 0.0822&\underline{0.1586} $\pm$ 0.0455 &0.6154 $\pm$ 0.0814 & 0.2337 $\pm$ 0.0533 &         &0.6021 $\pm$ 0.0478 &0.2142 $\pm$ 0.0353 & 0.6024 $\pm$ 0.0201& 0.3089 $\pm$ 0.0145 \\
TARNet &0.6659 $\pm$ 0.0812 & 0.2102 $\pm$ 0.0854 &0.7068 $\pm$ 0.0985 & 0.2065 $\pm$ 0.0798 &           &0.6071 $\pm$ 0.0544 &0.3021 $\pm$ 0.0343 & 0.5799 $\pm$ 0.0215& 0.3077 $\pm$ 0.0197 \\
CFRNet$_{\text{MMD}}$ &0.7586 $\pm$ 0.0786 & 0.1688 $\pm$ 0.0766 &\underline{0.7782} $\pm$ 0.0925 &0.3867 $\pm$ 0.0945 &         & \textbf{0.6823} $\pm$ 0.0327 &0.2785 $\pm$ 0.0373 &0.5438 $\pm$ 0.0278 & 0.2887 $\pm$ 0.0212\\
CFRNet$_{\text{WASS}}$ &\underline{0.7942} $\pm$ 0.0773& 0.3022 $\pm$ 0.0699 & 0.6973 $\pm$ 0.0622 & 0.1897 $\pm$ 0.0523&          & 0.6459 $\pm$ 0.0386 &0.3227 $\pm$ 0.0355 &0.5996 $\pm$ 0.0231 &0.3455 $\pm$ 0.0182\\
MEMENTO &0.7374 $\pm$ 0.0721&0.2989 $\pm$ 0.0642&0.7305 $\pm$ 0.0587&0.3575 $\pm$ 0.0844  &         & 0.6552 $\pm$ 0.0322 & \underline{0.1937} $\pm$ 0.0284 & \underline{0.6430} $\pm$ 0.0231&0.2544 $\pm$ 0.0155 \\
Hydranet &0.6908 $\pm$ 0.1120&0.2096 $\pm$ 0.0839 & 0.8277 $\pm$ 0.0742& 0.2587 $\pm$ 0.0918&            & 0.6088 $\pm$ 0.0321 &0.2912 $\pm$ 0.0401  &0.6202 $\pm$ 0.0133 &0.2517 $\pm$ 0.0221\\\midrule
M$^3$TN  &\textbf{0.8733} $\pm$ 0.0622& \textbf{0.1297} $\pm$ 0.0672&\textbf{0.8077} $\pm$ 0.0703& \textbf{0.1565} $\pm$ 0.0733&        &\underline{0.6804} $\pm$ 0.0298 &\textbf{0.1878} $\pm$ 0.0186 & \textbf{0.6906} $\pm$ 0.0211 & \textbf{0.1988} $\pm$ 0.0177
\\\bottomrule
\vspace{-0.5cm}
    \end{tabular}}
\end{table*}
\begin{table}[!t]
    \centering
    \vspace{-0.5cm}
    \caption{Results of the ablation studies on the Production dataset, where the best and second best results are marked in bold and underlined, respectively.}\label{tab:ablation}
  \resizebox{\linewidth}{!}{  \begin{tabular}{ccccc}\\ \toprule
\textbf{Methods} & mQini$\uparrow$ & sdQini$\downarrow$ & mKendall$\uparrow$ & sdKendall$\downarrow$ \\ \midrule
    M$^3$TN  (w/o MMoE) & \underline{0.6408} $\pm$ 0.0297&\underline{0.1899} $\pm$ 0.0178 & \underline{0.6403} $\pm$ 0.0201 & \underline{0.2366} $\pm$ 0.0191\\
    M$^3$TN  (w/o RM) & 0.6107 $\pm$ 0.0198 & 0.2120 $\pm$ 0.0243 & 0.6398 $\pm$ 0.0199 & 0.2708 $\pm$ 0.0259\\
    M$^3$TN  &\textbf{0.6804} $\pm$ 0.0298 &\textbf{0.1878} $\pm$ 0.0186 & \textbf{0.6906} $\pm$ 0.0211 & \textbf{0.1988} $\pm$ 0.0177\\
   \bottomrule
    \end{tabular}}
    \label{tab:my_label}
\end{table}
\label{sec:exp}
\subsection{Experimental Setup}
\noindent
\textbf{Datasets.}
We conduct experiments on a public dataset MultiTwins~\cite{almond2005costs} and a real-world industrial dataset Production to show the effectiveness of our method. The Production dataset is collected from one of the largest short video platforms in China. For such kind of short video platforms, video sharpening is an important user experience indicator. Different video sharpening degrees can lead to different user experiences. Through random experiments within two weeks, we conducted users with three degrees of sharpened videos ($T=1, 2, 3$) as the treatment group and users with normal videos ($T=0$) as the control group. The response is the total viewing time of users' short videos in a day; the resulting dataset contains more than 7 million users with 108 features describing user relative characteristics. 

\noindent
\textbf{Baselines.}
To evaluate the effectiveness of our M$^3$TN, we compare the performance of M$^3$TN with S-Learner~\cite{kunzel2019metalearners}, T-Learner~\cite{kunzel2019metalearners}, TARNet~\cite{shalit2017estimating}, CFRNet~\cite{shalit2017estimating}, MEMENTO~\cite{mondal2022memento} and Hydranet~\cite{velasco2022hydranet}. Especially for CFRNet, we compare the two variants with different Integral Probability Metrics (IPM), \textit{i.e.}. CFRNet$_{\text{MMD}}$ and CFRNet$_{\text{WASS}}$.

\noindent
\textbf{Evaluation Metrics.}
Following the setup of previous work~\cite{belbahri2021qini}, we employ two evaluation metrics commonly used in uplift modeling, \textit{i.e.}, the \textit{Qini coefficient}, and \textit{Kendall's uplift rank correlation}. Especially for multi-valued treatment settings, we use mQini and sdQini, the mean and the standard error of the Qini coefficient for all possible treatments, respectively. Similarly, we can also have mKendall and sdKendall. 

\noindent
\textbf{Implementation Details.}
We implement all the baselines and our M$^3$TN by Pytorch 1.10. We use the parameter search package Optuna~\cite{akiba2019optuna} to search the best parameters for all the baselines and our M$^3$TN; we use mini as the objective of the parameter tuning process. 

\subsection{Overall Performance (RQ1)}

We present the comparison results of MultiTwins and Production datasets in Table \ref{tab:overall}, and the reported results are the mean $\pm$ standardize over five different random seeds. Then, we have the following observations: For the MultiTwins dataset, 1) T-Learner outperforms S-Learner; even though some baselines use more complex architectures, the performance of T-Learner is also competitive on some metrics. 
2) By designing more complex architecture, TARNet, CFRNet, MEMENTO, and Hydranet can get better performance, But the improvement is not obvious on some metrics. 
This means that the architecture change with naively extending the binary treatment model structure may not yield much gain without considering the effectiveness and efficiency. 
3) Unlike other baselines, our M$^3$TN consistently outperforms all baselines in most cases. 
Since we use mQini as the hyper-parameter tuning objective, we significantly improve mQini, even though there may be some fluctuations for other metrics.
For the Production dataset, 1) S-Learner and T-Learner are still relatively stable and have competitive results. 
2) For the baselines, naively extending the binary treatment architecture suffers from a performance bottleneck, where the numerous model parameters may cause learning shocks and make the model difficult to be optimized with many different treatment groups.
3) Our M$^3$TN considers the structure effectiveness and the model efficiency. Then, our M$^3$TN still retains the best performance over all the baselines, which further shows the rationality of the proposed architecture. 
Combined with the experiments, one public dataset and one production dataset, we validate the effectiveness of our M$^3$TN; the results show that the architecture of our M$^3$TN can obtain better performance in the uplift modeling with multi-valued treatment.

\subsection{Ablation Studies (RQ2)}
We conduct the ablation studies of our M$^3$TN, and we analyze the role played by each module. 
We sequentially remove the two components of the M$^3$TN, \textit{i.e.} the Multi-gate Mixture-of-Experts (MMoE) of the feature representation module and the reparameterization module (RM). 
Then, construct two variants of M$^3$TN, which are denoted as M$^3$TN (w/o MMoE) and  M$^3$TN (w/o RM). 
We present the results in Table \ref{tab:ablation}, and the reported results are the mean $\pm$ standard deviation over five different random seeds. 
From the results, removing any part may bring performance degradation. 
Moreover, removing the RM brings more performance degradation than MMoE, which shows that the RM can reduce the cumulative errors in our problem setting.
All the modules are helpful in improving the effectiveness of our M$^3$TN. 

\subsection{Complexity Evaluation (RQ3)}
\begin{figure}[!t]
    \centering
    \includegraphics[width=0.55\linewidth]{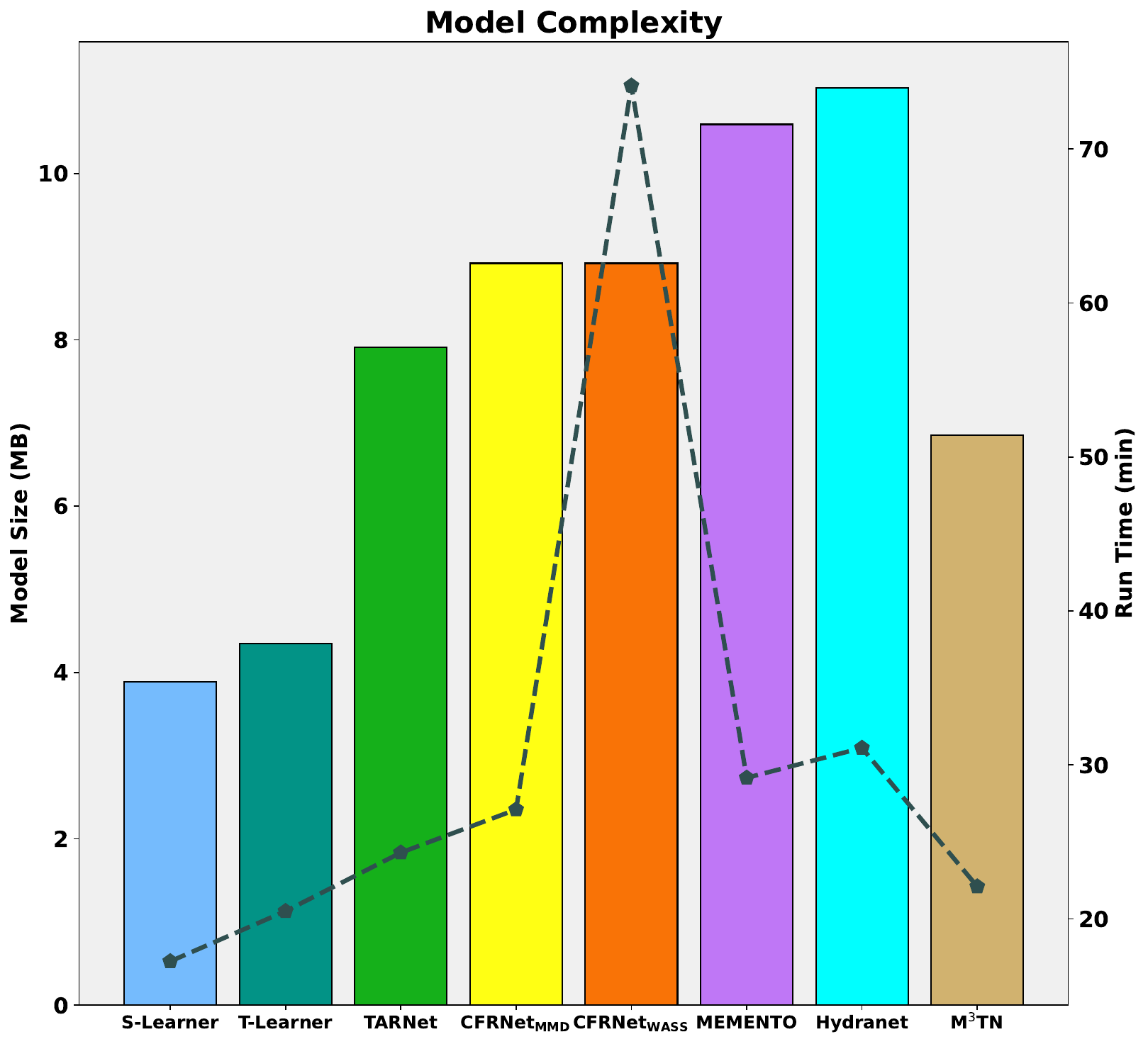}
    \caption{Model complexity (\textit{i.e.} model size and run time) comparison of all the baselines and our M$^3$TN.}
\vspace{-0.5cm}
    \label{fig:model_size}
\end{figure}
To compare the model efficiency, we evaluate the model complexity between our M$^3$TN and other baselines. 
To achieve a fair comparison, for the methods with shared bottom feature representation, we keep the same network structure of this module between them (\textit{i.e.} TARNet, CFRNet, MEMENTO, and Hydranet). 
We report the model size and run time in Fig.~\ref{fig:model_size}.
In particular, the run time is obtained for 20 training epochs on the Production dataset; we do not include the evaluation time cost, and all the experiments are conducted on NVIDIA V100 GPUs. 
Unsurprisingly, due to the design of feature representation with a Multi-gate mixture of experts, the model size of our M$^3$TN is smaller than TARNet, CFRNet, MEMENTO, and Hydranet.
For the run time, our M$^3$TN is less than MEMENTO, Hydranet and CFRNet and competitive to TARNet. Especially, due to the high computation cost of wasserstein distance~\cite{vallender1974calculation}, the CFRNet$_{\text{WASS}}$ shows the maximum run time among all the methods. 
Hydrant has the second longest run time because of the complex target regularization loss.
Thus, our M$^3$TN has a smaller model size, lower run time, and better performance, which is an effective and efficient uplift model.

\subsection{Hyperparameter Sensitivity (RQ4)}
\begin{figure}[!t]
    \centering
    \includegraphics[width=0.55\linewidth]{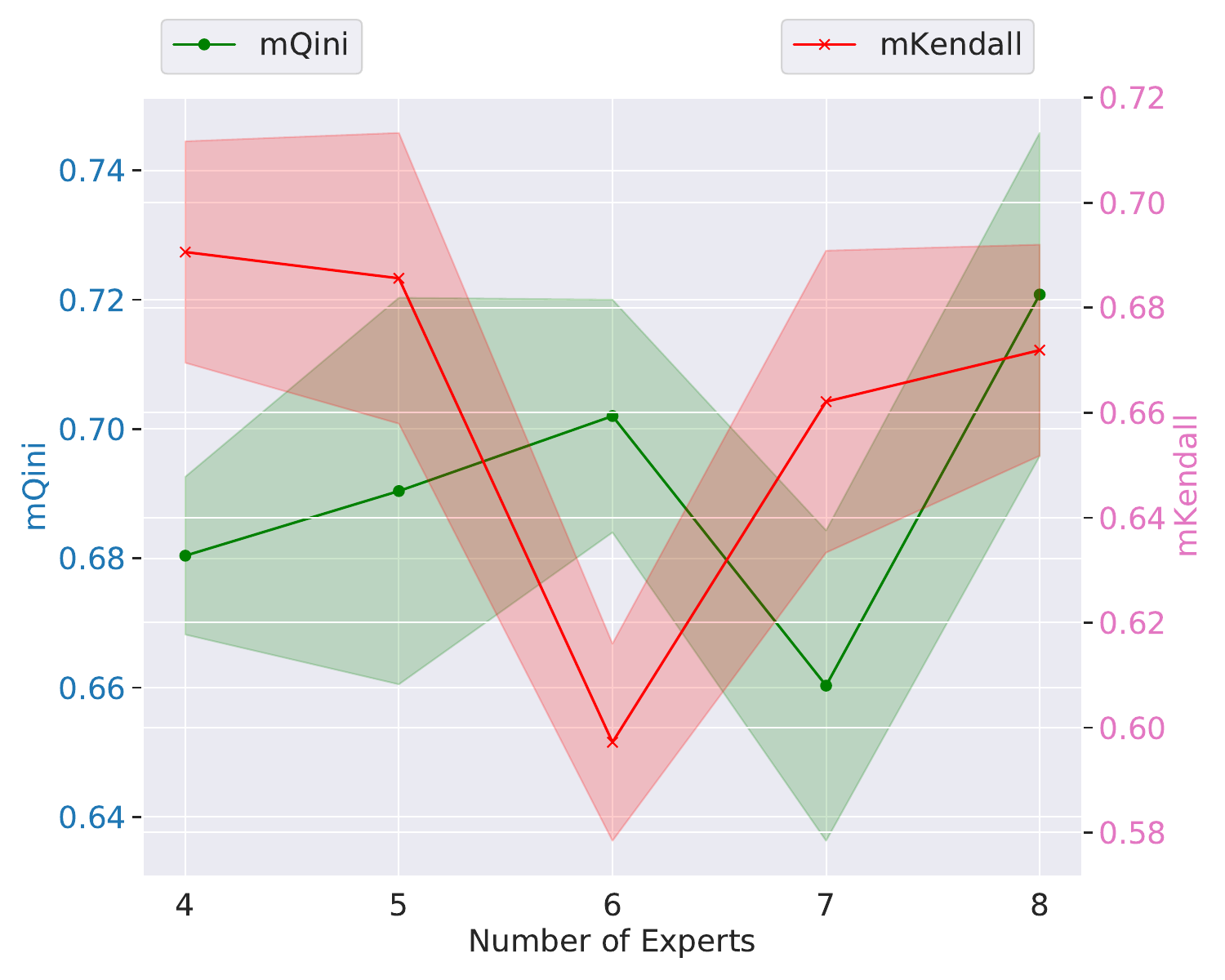}
    \caption{The number of experts sensitivity analysis results of our M$^3$TN.}\label{fig:hyper}
\end{figure}

We conduct the sensitivity analysis to the essential hyperparameter: the number of experts in the feature representation network. And we present the results in Fig.\ref{fig:hyper}.
From the results, we can see that our M$^3$TN is relative robust to the number of experts. 
However, to obtain the best performance among the mQini and mKendall, we need to adjust this hyperparameter carefully.

\section{Conclusion}
\label{sec:conclusion}

In this paper, to improve the effectiveness and efficiency of multi-valued treatment uplift modeling, we propose a novel M$^3$TN. Our M$^3$TN consists of two components: 1) a feature representation module with Multi-gate Mixture-of-Experts to improve the efficiency; 
2) a reparameterization module with modeling uplift explicitly to improve the effectiveness. 
Also, we conduct extended experiments on a public dataset and a real-world production dataset, which show the effectiveness and efficiency of our M$^3$TN. In future work, we aim to adapt theoretical results from multi-task learning to provide more theoretical analysis of multi-valued treatment uplift models.

\section{Acknowledgement}

This work is supported in part by National Natural Science Foundation of China (No. 62102420), Beijing Outstanding Young Scientist Program NO. BJJWZYJH012019100020098, Intelligent Social Governance Platform, Major Innovation \& Planning Interdisciplinary Platform for the ``Double-First Class'' Initiative, Renmin University of China, Public Computing Cloud, Renmin University of China, fund for building world-class universities (disciplines) of Renmin University of China, Intelligent Social Governance Platform.
\bibliographystyle{IEEEbib}
\bibliography{strings,refs}

\end{document}